\documentclass[11pt,a4paper]{article}
\usepackage[hyperref]{AACL-IJCNLP2020}
\usepackage{tabularx,booktabs}
\usepackage{times}
\usepackage{latexsym}
\usepackage{graphicx}
\usepackage{todonotes}

\usepackage{microtype}

\aclfinalcopy % Uncomment this line for the final submission

\title{Measuring Linguistic Diversity During COVID-19}

\author{Jonathan Dunn \\
  Department of Linguistics \\
  University of Canterbury \\
  Christchurch, New Zealand \\
  \small{\texttt{jonathan.dunn@canterbury.ac.nz}}\\\And
  Tom Coupe \\
  Department of Economics \\
  University of Canterbury \\
  Christchurch, New Zealand \\
  \small{\texttt{tom.coupe@canterbury.ac.nz}}\\\AND
  Benjamin Adams \\
  Department of Computer Science and Software Engineering \\
  University of Canterbury \\
  Christchurch, New Zealand \\
  \small{\texttt{benjamin.adams@canterbury.ac.nz}}\\
  }

\date{}

\begin{document}
\maketitle
\begin{abstract}
Computational measures of linguistic diversity help us understand the linguistic landscape using digital language data. The contribution of this paper is to calibrate measures of linguistic diversity using restrictions on international travel resulting from the COVID-19 pandemic. Previous work has mapped the distribution of languages using geo-referenced social media and web data. The goal, however, has been to describe these corpora themselves rather than to make inferences about underlying populations. This paper shows that a difference-in-differences method based on the Herfindahl-Hirschman Index can identify the bias in digital corpora that is introduced by non-local populations. These methods tell us \textit{where} significant changes have taken place and whether this leads to increased or decreased diversity. This is an important step in aligning digital corpora like social media with the real-world populations that have produced them.

\end{abstract}

\section{Biases in digital language data}

Data from social media and web-crawled sources has been used to map the distribution of both languages \cite{mocanu2013twitter,gonccalves2014crowdsourcing,lamanna2018immigrant,Dunn2020} and dialects \cite{eosx14,Cook2017,Dunn2019a,Dunn2019,Grieve2019}. This line of research is important because traditional methods have relied on census data and missionary reports \cite{Eberhard2020, Board2020}, both of which are often out-of-date and can be inconsistent across countries. At the same time, we know that digital data sets do not necessarily reflect the underlying linguistic diversity in a country: the actual population of South Africa, for example, is not accurately represented by tweets from South Africa \cite{DunnAdams2019}. 

This becomes an important problem as soon as we try to use computational linguistics to tell us about \textit{people} or \textit{language}. For example, if an application is using Twitter to track sentiment about COVID-19, that tracking is meaningless without good information about how well it represents the population. Or, if an application is using Twitter to study lexical choices, that study depends on a relationship between lexical choices on Twitter and lexical choices more generally. In other words, the more we use digital corpora for scientific purposes, the more we need to control for \textit{bias} in that data. There are four sources of diversity-related bias that we need to take into account.

First, \textit{production bias} occurs when one location (like the US) produces so much digital data that most corpora over-represent that location \cite{Jurgens2017}. For example, by default a corpus of English from the web or Twitter will mostly represent the US and the UK \cite{kulshrestha2012geographic}. It has been shown that this type of bias can be corrected using population-based sampling \cite{DunnAdams2020} to enforce the representation of all relevant populations.

\begin{figure*}
  \includegraphics[width=460pt]{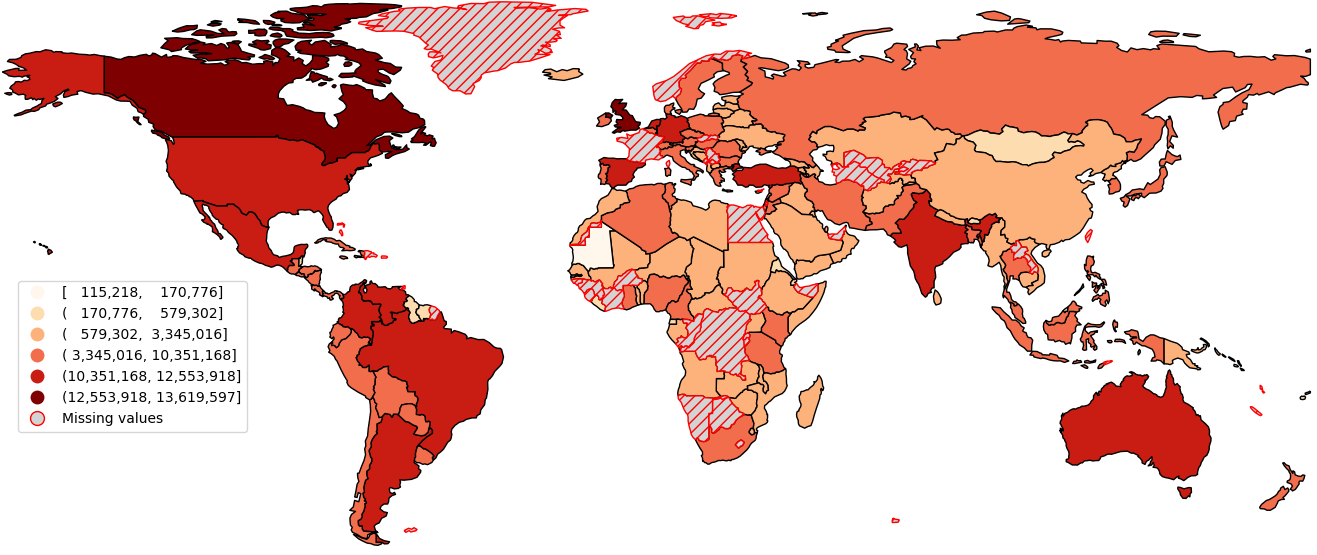}
  \caption{Number of observations per country.}
  \label{fig:map1}
\end{figure*}

Second, \textit{sampling bias} occurs when a subset of the population produces a disproportionate amount of the overall data. This type of bias has been shown to be closely related to economic measures: more wealthy populations produce more digital language per capita \cite{DunnAdams2019}. By default, a corpus will contain more samples representing wealthier members of the population. Thus, this is similar to production bias, but with a demographic rather than a geographic scope.

Third, \textit{non-local bias} is the problem of over-representing those people \textit{in} a place who are not \textit{from} that place: tourists, aid workers, students, short-term visitors, etc. For example, in countries with low per-capita GDP (i.e., where local populations often lack internet access) digital language data is likely to represent outsiders like aid workers. On the other hand, in countries with large numbers of international tourists (e.g., New Zealand), data sets are likely to instead be contaminated with samples from these tourists.

Fourth, \textit{majority language bias} occurs when a multi-lingual population only uses some of its languages in digital contexts \cite{lackaff2016local}. Most often, majority languages like English and French are used online while minority languages are used in face-to-face contexts. The result is that even though an individual may be represented in a corpus, the full range of their linguistic behaviours is \textit{not} represented. This is the only type of bias not quantified in this paper. For example, it is possible that changes in linguistic diversity are caused by a shift in behaviour, rather than a shift in population characteristics.

Of the three sources of bias that we examine here, non-local bias is the most difficult to uncover \cite{ghg14, johnson2016geography}. We can identify production bias when the amount of data per country exceeds that country's share of the global population. In this sense, the ideal corpus of English would equally represent each country according to the number of English speakers in that country. Within a country, we can measure the amount of sampling bias by looking at how economic measures like GDP and rates of internet access correspond with the amount of data per person. Thus, we could use median income by zip code to ensure that the US is properly represented. But non-local bias is more challenging because we need to know which samples from a place like New Zealand come from those speakers who are only passing through for a short time.

Only with widespread restrictions on international travel during the COVID-19 pandemic do we have access to a collection of digital language from which non-local populations are largely absent \cite{gossling2020pandemics, hale2020variation}. This paper uses changes in linguistic diversity during these travel restrictions, against a historical baseline, to calibrate computational measures that support language and population mapping. This is a part of the larger problem of estimating population characteristics from digital language data.

We start by describing the data used for the experiments in the paper (Section 2), drawn from Twitter over a two-year period. We then explore sources of bias in this data set by looking at production bias and sampling bias (in Section 3) and then developing a baseline of temporal variation in the data (in Section 4). We introduce a measure of geographic linguistic diversity (Section 5). Then we use this measure to find which countries and languages are most contaminated by non-local populations (in Section 6). Finally, we examine the results to find where the linguistic landscape has changed during the COVID-19 pandemic.

\section{Data sources}

\begin{table}
\centering
\begin{tabular}{|l|r|r|r|}
\hline
\textbf{Region} & \textbf{N.} & \textbf{Pop} & \textbf{Data} \\
\hline
Africa, Southern & 12.28m & 1.0\% & 2.0\% \\
Africa, Sub & 43.87m & 10.1\% & 7.0\% \\
Africa, North & 16.60m & 3.4\% & 2.7\% \\
\hline
America, Brazil & 10.96m & 2.8\% & 1.8\% \\
America, Central & 66.12m & 2.9\% & 10.6\% \\
America, North & 24.64m & 4.8\% & 4.0\% \\
America, South & 77.79m & 2.9\% & 12.5\% \\
\hline
Asia, East & 15.88m & 22.3\% & 2.6\% \\
Asia, Central & 15.08m & 2.7\% & 2.4\% \\
Asia, South & 30.06m & 23.3\% & 4.8\% \\
Asia, Southeast & 31.88m & 8.4\% & 5.1\% \\
\hline
Europe, East & 51.48m & 2.4\% & 8.3\% \\
Europe, Russia & 9.38m & 2.0\% & 1.5\% \\
Europe, West & 155.74m & 5.7\% & 25.0\% \\
\hline
Middle East & 36.58m & 4.5\% & 5.9\% \\
Oceania & 24.92m & 0.8\% & 4.0\% \\
\hline
\textbf{Total} & \textbf{623.33m} & \textbf{100\%} & \textbf{100\%} \\
\hline
\end{tabular}
\caption{Distribution of data by region.}
\label{tab:exp1}
\end{table}

We draw on Twitter data sampled globally from 10k cities over a 25-month period (July 2018 through August 2020). This city-based collection reduces production bias from the start (as opposed to collecting data by user or search term) because it forces non-central cities to be included. The cities are selected to represent the global population and all retweets are removed. This provides 623 million tweets, distributed across regions as shown in Table~\ref{tab:exp1} with each region's share of the data and of the world's population. 

This table provides a clear illustration of production bias. East Asia, for example, accounts for 22.3\% of the world's population but only 2.6\% of the data. We see the reverse in Western Europe, which provides 25\% of the data but only 5.7\% of the population. Population-based sampling is an effective method for correcting this bias \cite{DunnAdams2020}, if the goal is to produce a corpus representing the actual distribution of speakers. Our goal here is to find which countries contain data from non-local populations. To do this, we need to find out if the data has a stable geographic distribution that is driven by the underlying population.

The idNet language identification package is used to provide language labels \cite{Dunn2020}. Any tweet under 40 characters (after cleaning URLs and hashtags) is removed because of reduced identification accuracy below this threshold.  The average tweets per month per country is visualized in Figure~\ref{fig:map1}. Because we are looking at change over time by country, the data is binned into potentially small categories (e.g., Nigeria in July 2019). Both the table and the map show that countries in East Asia are under-represented. Thus, we use significance testing \textit{within} countries when looking for change over time.

\begin{figure*}[t]
  \includegraphics[width=460pt]{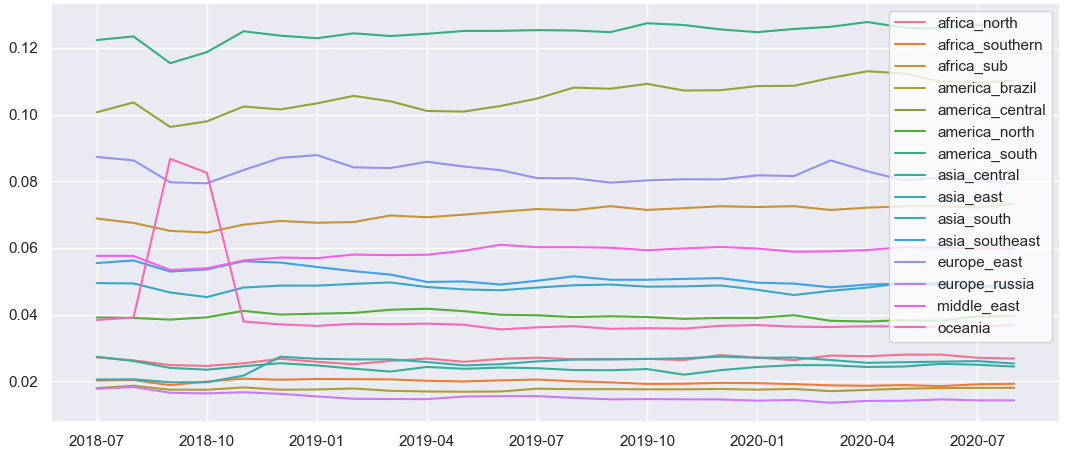}
  \caption{Geographic distribution of data by region by month.}
  \label{fig:map2}
\end{figure*}

\section{Demographics and language use}

\begin{figure}
  \includegraphics[width=220pt]{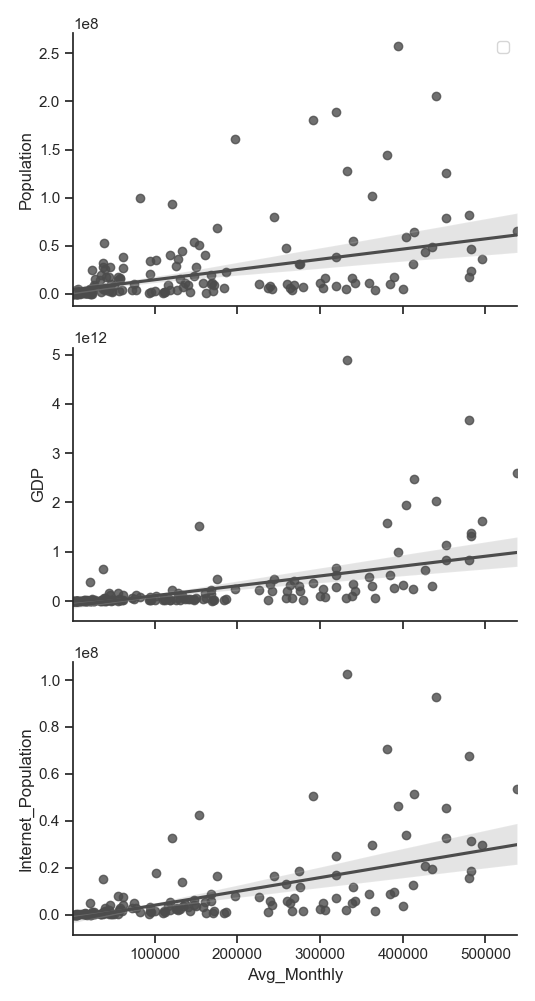}
  \caption{Relationship between data and demographic factors: \textit{Population}, \textit{Internet Access}, and \textit{GDP} (With Outliers Removed).}
  \label{fig:reg11}
\end{figure}

The next question is the degree to which the production of this data is driven by underlying populations (potential production bias) and by demographic factors like GDP (potential selection bias). We start, in Figure~\ref{fig:reg11}, by looking at the relationship between each country's population and share of the corpus. This expands on the region aggregations in Table~\ref{tab:exp1} by dividing regions into countries. Each country is an observation that is represented by its average monthly data production and several demographic factors. Overall, there is a very significant correlation (Pearson) between population and the amount of data from each country (0.46). Thus, the number of people in a country is an important factor explaining how much data that country produces. While this is significant, however, it also means that there are many other factors that influence the geographic distribution of the data.

\begin{figure*}[t]
  \includegraphics[width=440pt]{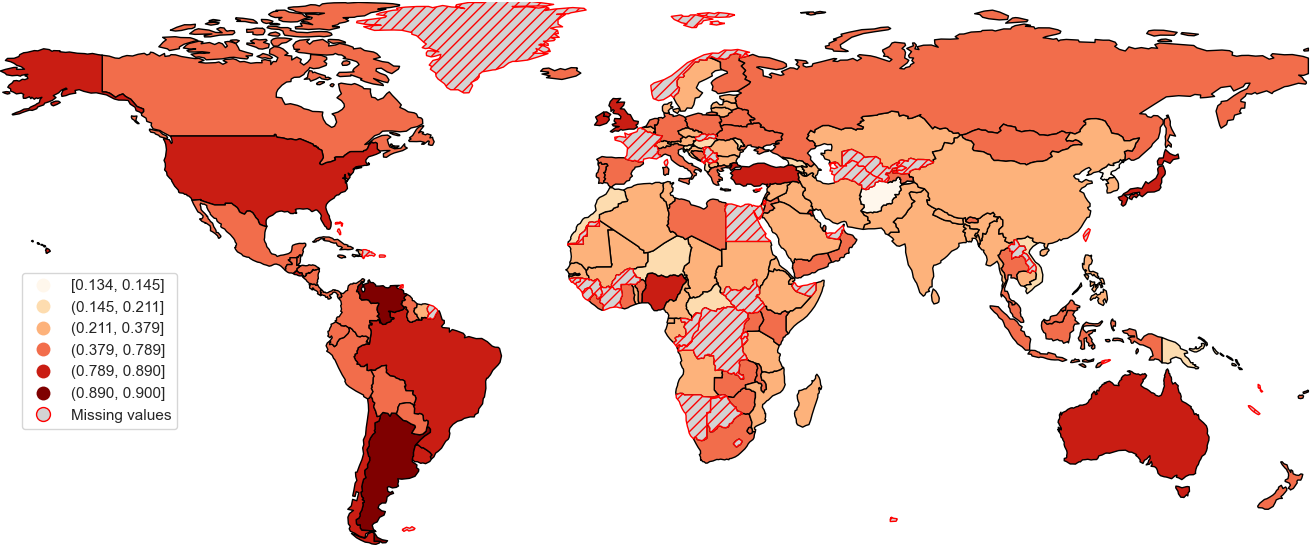}
  \caption{Herfindahl-Hirschman Index of the distribution of languages by country.}
  \label{fig:map3}
\end{figure*}

To better understand the factors influencing the geographic distribution of the data, we work with three variables: \textit{population}, the number of people in each country; \textit{internet population}, the number of internet users in each country; and \textit{GDP}, a measure of each country's economic output \cite{UnitedNations2011,UnitedNations2017,UnitedNations2017a}. Figure~\ref{fig:reg11} shows three regression plots in which these variables (on the y axis) are compared with the average monthly data production per country (given in number of tweets per month on the x axis). 

In each case, there is a close relationship between data production and demographics, with several extreme outliers. For \textit{population}, the outliers are China and India. Both are highly populated countries with significantly lower than expected data production (especially China). Both countries have relatively low rates of internet access: 38\% for China and 11\% for India; this lowers the total population in each country. Thus, although the populations are quite large, most of the population is not able to produce digital language data. For the influence of GDP, the outliers are the US and China. For the US, in particular, the GDP is quite high: there seems to be a ceiling after which increased GDP is unlikely to influence digital behaviours. Further, that GDP is not evenly distributed across the entire population. For the influence of internet access, the outliers are again China and the US. With a few notable exceptions there is a relatively close relationship between data production and the demographic factors of each country.

With these three outliers removed (the US, China, India), there are very significant correlations between these three variables and the geographic distribution of the data: 0.46 (population), 0.61 (population with internet access), and 0.59 (GDP). This leaves some unexplained production factors. The most obvious missing factor here is social media platforms specific to given countries (e.g., Sina Weibo). These alternative platforms will siphon away enough users to distort the representation of a population given access only to other platforms. Further, Twitter is banned in China: because only some companies are allowed to use it through specific VPNs, the text is not representative of language use in China. Casual users of Twitter will use a VPN through another country which would distort this method of data collection.

Regardless, this section has shown that we can explain a significant portion of the geographic distribution of the data. This is important because we want to describe \textit{populations} by observing \textit{digital corpora}. If there is no relationship between the two in terms of distribution, it is difficult to make such inferences. What we have seen, however, is that there is a very significant relationship. What is the required threshold for establishing a relationship like this? We should think about this as a metric for evaluating digital corpora: data with a stronger relationship to demographic variables are more representative. The next question is whether this relationship remains stable over time: can we depend on these demographic factors across the entire period?

\section{Controlling for temporal variation}

The next question is whether these production factors are stable over time. Here we build a baseline for temporal variation: to what degree is the data subject to unrelated fluctuations that will reduce our ability to assign a cause-and-effect relationship to linguistic diversity during travel restrictions?

Although the same collection and processing methods are maintained over the two-year period, there is variation in the total number of observations (tweets) per month. There are many reasons why this might be the case. What matters to us, though, is the relative share of each country. In other words, the population does not change from month to month in the same way that the number of tweets changes. Regardless of the total amount of data collected per month, is the geographic distribution consistent? Figure~\ref{fig:map2} shows stability over time by representing the relative proportion of observations per region by month. Western Europe is removed for the sake of clarity, as it represents a significantly higher share (roughly 25\%). The distribution of samples is consistent over time. The main exception is that, for a two-month period in 2018, there is much more data from Oceania. 

We use a t-test to find out if the share of each region is stable over time. If the distribution changes significantly, then it may be hard to determine the cause of any individual change. None of the regions show a significant fluctuation; this is helpful because it shows that there is not random noise in the data that could interfere with measures of linguistic diversity. The difference-in-differences methods we use in Section 8 would control for such noise, but this gives us further confidence. We use a t-test, rather than a time-specific test like Dickey-Fuller, because we are interested in consistency rather than in non-stationarity. These results show that, in the aggregate, the distribution of samples remains constant. But how much variation within individual countries does this region-based measure disguise? To answer this, we look at the same t-test approach by country: do individual countries vary widely in their relative production? No countries show a significant change.

\begin{figure*}[t]
  \includegraphics[width=440pt]{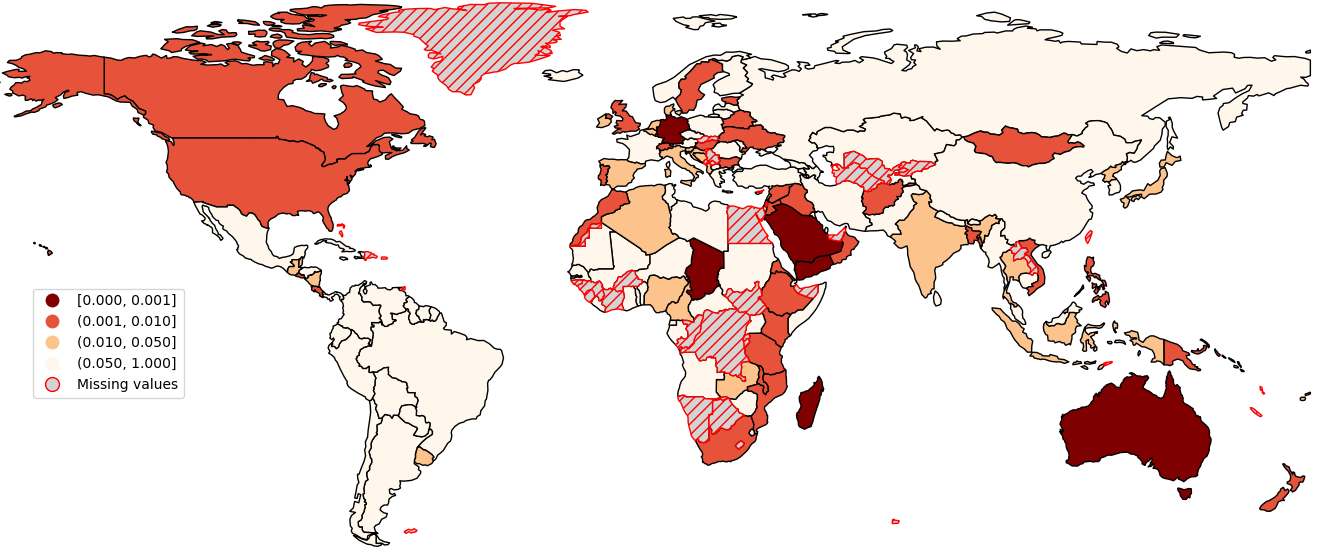}
  \caption{Countries with significant change in linguistic diversity during travel restrictions.}
  \label{fig:map4}
\end{figure*}

These findings show that we can largely focus on diversity,  the distribution of languages within a country by month, rather than on the production of data over month as in Section 3. There is natural variation in the data, of course, and this is taken into account in our later approaches. For example, if we compute the correlation between population and language production (as in Section 3) for each month in isolation, there is no significant difference over time. This stability is important for creating a baseline against which to understand demographic changes during travel restrictions. Because the relationship between demographics and the data set remains stable, we can focus specifically on changes in linguistic diversity. 

\section{Measuring linguistic diversity}

Linguistic diversity is an important part of accurate language and population mapping. The goal is to have a single measure that can tell us how much language contact is taking place and which communities are multi-lingual. To do this we must generalize across specific languages: linguistic diversity in the US might involve English and Spanish, but it might involve Portuguese and Spanish in Brazil.

We measure linguistic diversity as a probability distribution over languages for each country. Drawing on previous work on short-sample language identification, this paper includes 464 languages across 157 countries. For each country, then, we have a relatively accurate identification of which languages are used on Twitter. Given this probability distribution for each country, we compare countries using the Herfindahl-Hirschman Index (HHI) as shown in Figure ~\ref{fig:map3}. The HHI was developed in economics to measure market concentration: the more of a given industry is dominated by a small number of companies, the higher the HHI \cite{hirschman1980national}. The measure is derived using the sum of the square of shares, in this case the share of each language in each country. The higher the HHI (the darker red) for a country, the more one language dominates the linguistic landscape.

\begin{table}
\centering
\begin{tabular}{|c|r|r|r|}
\hline
~ & \textbf{ISR} & \textbf{IND} & \textbf{USA} \\
\hline
HHI & 0.207 & 0.356 & 0.852 \\
\hline
L1 & 27.3\% & 50.8\% & 92.3\% \\
L2 & 25.9\% & 30.8\% & 2.6\% \\
L3 & 23.5\% & 3.4\% & 0.6\% \\
L4 & 7.5\% & 2.5\% & 0.6\% \\
L5 & 5.3\% & 1.4\% & 0.4\% \\
\hline
\end{tabular}
\caption{Sample language distributions by country.}
\label{tab:exp2}
\end{table}

Thus, the HHI is higher when the distribution is centered around just a few languages. For example, in Table~\ref{tab:exp2} we focus on three countries that show a range of linguistic diversity: Israel, India, and the US. Israel has the lowest HHI (0.207). Looking at the share of the top five languages, we see roughly equal usage of three languages (in the 20s) followed by two significant minority languages. This lower HHI reflects the fact that a number of languages are being used together: no language has a monopoly. On the other extreme, the US has one of the highest values for HHI (0.852). There is one very dominant language (92\%), one significant minority language (2.6\%), and a number of very insignificant languages. English has a metaphoric monopoly on the linguistic landscape of the US.

\begin{figure*}[t]
  \includegraphics[width=440pt]{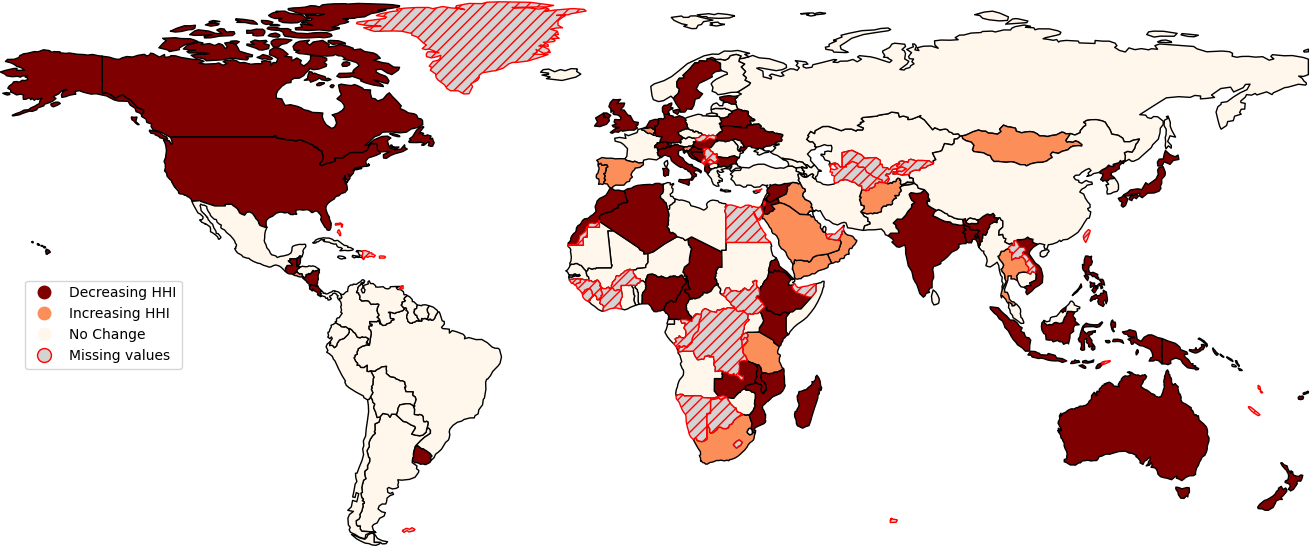}
  \caption{Increasing vs. decreasing HHI during travel restrictions.}
  \label{fig:map5}
\end{figure*}

Figure~\ref{fig:map3} shows linguistic diversity across the world: lighter countries (like Israel) have a mix of languages while darker countries (like the US) are mostly monolingual. There are many linguistic landscapes around the world, ranging from multilingual to monolingual. This Figure~\ref{fig:map3} is a baseline representation, averaged across the entire time period (July 2018 to August 2020). It is possible that this averaged representation disguises temporal fluctuations. We have already seen that there are only a few changes in the share of data per country per month, and no significant change in the relationship between the data set and demographic factors like GDP. The question here is whether there is arbitrary variation in the linguistic diversity per country per month. In other words, if Israel becomes significantly more diverse every three months, it will be difficult to find out what is causing those changes. We use a a t-test for the mean of each country to determine if each country's diversity is actually a single group. There are no significant fluctuations across the period as a whole.

\section{Finding non-local populations}

To what degree do countries change during travel restrictions resulting from COVID-19? We have a measure of diversity (the HHI) and data collected by month. The basic approach is to create two groups of samples: first, months during the pandemic (March through August, 2020); second, months not during the pandemic (March through August, 2019). These two groups are aligned by month so that seasonal fluctuations are taken into account (e.g., tourism high season in February for New Zealand and in July for Italy). Given these two groups of samples, we use a t-test for two independent samples to determine whether these groups are, in fact, different. If we reject the null hypothesis, it means that linguistic diversity during travel restrictions is significantly different than the seasonally-adjusted baseline.

The results show that 70 countries have a changed linguistic landscape during the pandemic. This is visualized in Figure~\ref{fig:map4}, with p-values classed into highly significant (under 0.001), very significant (under 0.01), and significant (under 0.05). We see, for example, that the US and Canada undergo significant change, but not Mexico and South America. There are clear geographic patterns in linguistic change: North but not Central or South America; East Africa but not West Africa; South/east Asia but not East Asia; Europe but not Russia. We will examine in more detail how and why the linguistic landscape changes in Section 7.

These significant changes during international travel restrictions show that our measure (the HHI) and our data (tweets) offer a meaningful representation of underlying populations. If the data did not represent populations, we would not see the relationships examined in Section 3. There are no random fluctuations in the distribution of the data across countries or in the distribution of languages within countries. At the same time, given a massive social change (i.e., the COVID-19 pandemic), the measure clearly identifies changes in the linguistic landscape. Thus, the measure is both precise (not disguised by noise) and accurate (observing change where we expect it). The key point is that the change in diversity during the COVID-19 period is identifiable against the background noise.

A country's linguistic landscape could change by becoming more diverse (i.e., with more languages) or by becoming less diverse (i.e., with fewer languages). Which is causing the significant changes that we are observing? Figure~\ref{fig:map5} distinguishes between countries with an increasing HHI (becoming more monolingual) and a decreasing HHI (becoming more multilingual). We can think about two contexts in which this change can take place: a country like India might look more multilingual because non-local tourists who speak English are no longer creating noise in the data; or, a country like South Africa might look more monolingual because its own English-speaking citizens abroad are returning home. The flow of international travellers changes the balance of locals and non-locals in both directions (leaving and coming home).

\section{Identifying out-of-place populations}

Our task now is to use these changes during travel restrictions to identify which populations are out-of-place in ordinary times. In other words, if India has decreasing English use during the pandemic period, then we know that English is over-represented in the country as a result of non-local populations. We find these languages by repeating the comparison of pandemic vs. normal periods per country per month, but now we look at the share of individual languages rather than the HHI (in countries with a significant change). We are only interested in languages which account for at least 1\% of a country's usage. Less commonly used languages may be changing significantly but have less influence on a country's overall linguistic landscape.

\begin{table}
\centering
\begin{tabular}{|l|c|c|}
\hline
\textbf{Country} & \textbf{Normal} & \textbf{COVID} \\
\hline
Eritrea & 63.16\% & 41.94\% \\
Samoa & 45.00\% & 30.18\% \\
Cabo Verde & 27.78\% & 16.63\% \\
Equatorial Guinea & 33.08\% & 24.40\% \\
Madagascar & 53.08\% & 44.87\% \\
Kiribati & 31.10\% & 23.56\% \\
Tanzania & 34.43\% & 27.35\% \\
Mongolia & 30.32\% & 23.52\% \\
Chad & 45.48\% & 39.71\% \\
Sao Tome & 12.57\% & 7.14\% \\
Yemen & 14.44\% & 9.20\% \\
\hline
\end{tabular}
\caption{Major reductions in English.}
\label{tab:change1}
\end{table}

We start by looking at countries where the use of English falls dramatically during the pandemic period, in Table~\ref{tab:change1}. These dramatic reductions suggest that much of the population represented on Twitter is non-local: there is a change from 63\% to 42\% in Eritrea and from 53\% to 44\% English use in Madagascar. If the local population was well-represented on Twitter, we would not see this dramatic reduction in an international language. Thus, here we see an example of how digital data is biased towards non-local populations in countries where the local population has reduced internet access.

\begin{table}
\centering
\begin{tabular}{|l|c|c|c|}
\hline
\textbf{Country} & \textbf{Language} & \textbf{Normal} & \textbf{COVID} \\
\hline
Belarus & Russian & 69.05\% & 66.13\% \\
Ukraine & Russian & 54.60\% & 50.06\% \\
Lithuania & Russian & 20.09\% & 15.72\% \\
Latvia & Russian & 10.43\% & 8.26\% \\
\hline
Algeria & Arabic & 51.56\% & 46.77\% \\
Morocco & Arabic & 33.75\% & 28.53\% \\
Israel & Arabic & 27.75\% & 26.08\% \\
Tunisia & Arabic & 24.24\% & 19.65\% \\
Bhutan & Arabic & 6.25\% & 2.55\% \\
Moldova & Arabic & 2.71\% & 0.79\% \\
\hline
\end{tabular}
\caption{Major reductions in Russian and Arabic.}
\label{tab:change2}
\end{table}

The influence of non-local populations returning home is shown for Russian and Arabic in Table~\ref{tab:change2}. We see a major reduction in the use of Russian in countries like Ukraine that have had a strong Russian influence (from 54\% to 50\%). In both Ukraine and Belarus, there are other social and political factors that could influence the shift, since much of the population is bilingual (e.g., bilingual speakers in Ukraine putting aside the use of Russian for political purposes). But we also see similar changes in the use of Arabic. In Algeria it falls from 51\% to 46\% and in Morocco from 33\% to 28\%. These countries do not have the same political factors as Ukraine and Belarus, thus providing a clearer example of the exodus of non-local populations.

\begin{table}
\centering
\begin{tabular}{|c|l|c|c|}
\hline
\textbf{Country} & \textbf{Language} & \textbf{Normal} & \textbf{COVID} \\
\hline
SAU & Arabic & 70.10\% & 81.87\% \\
SAU & English & 12.18\% & 7.35\% \\
SAU & Turkish & 4.34\% & 2.12\% \\
SAU & Greek & 2.55\% & 1.65\% \\
\hline
BEL & French & 28.64\% & 34.72\% \\
BEL & English & 31.01\% & 26.83\% \\
BEL & Dutch & 27.08\% & 25.12\% \\
BEL & German & 2.26\% & 1.93\% \\
BEL & Portuguese & 1.51\% & 1.68\% \\
\hline
\end{tabular}
\caption{Changing landscape in Saudi Arabia and Belgium.}
\label{tab:change3}
\end{table}

We get a different view by looking at the change of languages \textit{within} a country, as with Belgium and Saudi Arabia in Table~\ref{tab:change3}. In Saudi Arabia we see a rise in Arabic at the expense of English, Turkish, and Greek. This reflects the exodus of non-local tourists and workers; but it also likely reflects the return of Saudi Arabians from countries like Algeria and Morocco that is suggested by Table~\ref{tab:change2}. In Belgium, we see a rise in French at the expense of English, Dutch, German, and Portuguese. This is a reflection of a reduction in non-local tourists. 

\begin{table}
\centering
\begin{tabular}{|c|l|c|c|}
\hline
\textbf{Country} & \textbf{Language} & \textbf{Normal} & \textbf{COVID} \\
\hline
NZL & English & 86.26\% & 84.13\% \\
NZL & Spanish & 2.13\% & 3.37\% \\
NZL & Portuguese & 2.30\% & 2.82\% \\
NZL & Indonesian & 0.89\% & 1.27\% \\
\hline
AUS & English & 89.51\% & 87.45\% \\
AUS & Portuguese & 1.83\% & 2.52\% \\
AUS & Spanish & 1.52\% & 2.08\% \\
AUS & Japanese & 0.99\% & 1.32\% \\
\hline
\end{tabular}
\caption{Changing landscape in Oceania.}
\label{tab:change4}
\end{table}

However, we see the opposite effect of tourists leaving when we look at New Zealand and Australia, two countries which have had closed borders (Table~\ref{tab:change4}). Here there is a \textit{reduction} in English usage within English-majority countries that takes place when international tourists stop arriving. The situation here is that there are so many English-speaking tourists (i.e., from the US and UK) that local immigrant languages like Spanish and Portuguese (part of the long-term local population) are drowned out by non-local tourists using English. Another possible explanation is that immigrant populations are increasingly using Twitter to communicate with non-local populations (e.g., with family and friends in their previous country).

\section{Sources of Change}

This paper has shown that there is a significant change in the linguistic diversity of many countries \textit{during} the travel restrictions caused by COVID-19. But to what degree are these changes \textit{related} to the travel restrictions themselves? For example, we could imagine a population that is changing over time which we just happen to observe in mid-change. It could be the case that a country has been becoming less diverse over the past decade because of fewer incoming immigrants; the approach taken so far in this paper would misinterpret such macro-trends to be a direct result of COVID-19. 

We use a difference-in-differences method \cite{Card94} to correct for this. The basic idea behind a difference-in-differences approach is to conduct a \textit{natural experiment} with a control group (here, data from 2018) and an effect group (here, data from 2020) differentiated by time. We have three months (July, August, September) that are shared across 2018, 2019, and 2020. So, using the same methods described above, we find out which countries have a significant change between 2019 and 2020. This is the period that takes place during travel restrictions. If travel restrictions influence linguistic diversity, we would expect such influence to take place during this period. We then find out if the countries which show a significant change in 2020 also show a significant change from 2018 to 2019. This provides a baseline: removing any country whose linguistic diversity was already in the process of changing.

Over this three-month period (July through September), 58 countries show a change in linguistic diversity during the pandemic. This is a smaller number than the main results reported above for two reasons: (i) the time span is shorter, giving less robust results and (ii) this particular time span came after some travel had resumed. Of these 58 countries that show a significant change in diversity, most (38) show no difference at all in the baseline period before the pandemic. Another eight show a much greater difference during the COVID-19 period (e.g., p-values of 0.03 vs 0.004 for baseline and COVID-19, respectively). This means that the pandemic has either created or has significantly contributed to 79.3\% of the cases of changing linguistic diversity. The remaining 20.7\% of changes, then, must have been created by macro-trends like immigration or changes in bilingual behaviour. The main conclusion from this difference-in-differences examination, however, is that most of these changes can be specifically connected to COVID-19.

\section{Conclusions}

The goal of this paper is to validate measures of linguistic diversity using changes in underlying populations during the COVID-19 pandemic. We have shown that there is a significant relationship between our data and the underlying population. Thus, what we are observing (tweets) can tell us about the people we want to study. At the same time, both the distribution of the data across countries and the distribution of languages within countries are stable. Thus, the data does not have random fluctuations that will get in the way. Using the HHI as a measure of diversity, there is a significant change in the linguistic landscape of 70 countries against a seasonally-adjusted baseline. This reflects non-local populations (e.g., the impact of tourists leaving a country or short-term visitors returning to their own countries). These results validate a measure of linguistic diversity that is based on digital language data and shows that we can correct for the bias introduced by non-local populations.

\bibliography{aacl-ijcnlp2020}
\bibliographystyle{acl_natbib}

\end{document}